\documentclass[sigconf, screen, preprint]{acmart}
\AtBeginDocument{%
  }

\setcopyright{acmlicensed}
\copyrightyear{2026}
\acmYear{2026}
\acmDOI{XXXXXXX.XXXXXXX}
\acmConference[ACMMM 2026]{the 34th ACM International Conference on Multimedia}{}{Rio de Janeiro, Brazil}
\acmISBN{978-1-4503-XXXX-X/2018/06}

\renewcommand\footnotetextcopyrightpermission[1]{}
\acmSubmissionID{3049}

\usepackage{amsmath, arydshln, pifont, multirow, booktabs, graphicx}

\usepackage{amssymb}
\usepackage[table]{xcolor}

\begin{document}

\title{DL-SLAM: Enabling High-Fidelity Gaussian Splatting SLAM in Dynamic Environments based on Dual-Level Probability}

%\email{webmaster@marysville-ohio.com}
%\affiliation{%
%  \institution{Institute for Clarity in Documentation}
%  \country{USA}
%}
%
%\author{Lars Th{\o}rv{\"a}ld}
%\affiliation{%
%  \institution{The Th{\o}rv{\"a}ld Group}
%\email{larst@affiliation.org}
%
%\author{Ziheng Xu}
%\affiliation{%
%  \institution{Beihang University}
%  \city{Beijing}
%  \country{China}
%}
%
%\author{Qingfeng Li}
%\affiliation{%
%	\institution{Beihang University}
%	\city{Beijing}
%	\country{China}
%}
%
%\author{Xuefeng Liu}
%\affiliation{%
%	\institution{Beihang University}
%	\city{Beijing}
%	\country{China}
%}
%
%\author{Chen Chen}
%\affiliation{%
%	\institution{Hangzhou Innovation Institute of Beihang
%		University}
%	\city{Hangzhou}
%	\country{China}
%}
%
%\author{Jianwei Niu}
%\affiliation{%
%	\institution{Beihang University}
%	\city{Beijing}
%	\country{China}
%}

\author{Ziheng Xu, Qingfeng Li, Xuefeng Liu, Chen Chen, Jianwei Niu}

\renewcommand{\shortauthors}{Xu et al.}

\begin{abstract}
  Recent advances in 3D Gaussian Splatting (3DGS) have enabled significant progress in dense dynamic Simultaneous Localization And Mapping (SLAM). Prevailing methods typically discard predefined dynamic objects, ignoring that transiently static objects offer valuable geometric constraints for pose estimation. A recent work attempts to leverage this potential by employing per-pixel uncertainty maps to quantify the magnitude of motion. While this approach enables transiently static objects to enhance pose estimation, it erroneously integrates these objects into the static map, resulting in persistent artifacts. Moreover, its reliance on purely geometric information leads to ambiguous object boundaries in the uncertainty maps. To overcome these limitations, we present DL-SLAM, a monocular Gaussian Splatting SLAM system built upon a novel dual-level probabilistic framework. Our method computes dynamic probability maps by combining semantic and geometric information. These pixel-level probabilities are lifted to 3D and aggregated to derive an object-level dynamic probability for each instance. Object-level probability enables the categorical pruning of dynamic Gaussians, resulting in an artifact-free static map. The static map, in turn, provides a geometrically consistent guidance to refine the pixel-wise probabilities, enhancing their reliability. Experimental results demonstrate that DL-SLAM outperforms existing approaches, improving tracking accuracy by up to 13\% while generating high-fidelity semantic maps.
\end{abstract}




\begin{teaserfigure}
	\centering
	\includegraphics[width=0.98\linewidth]{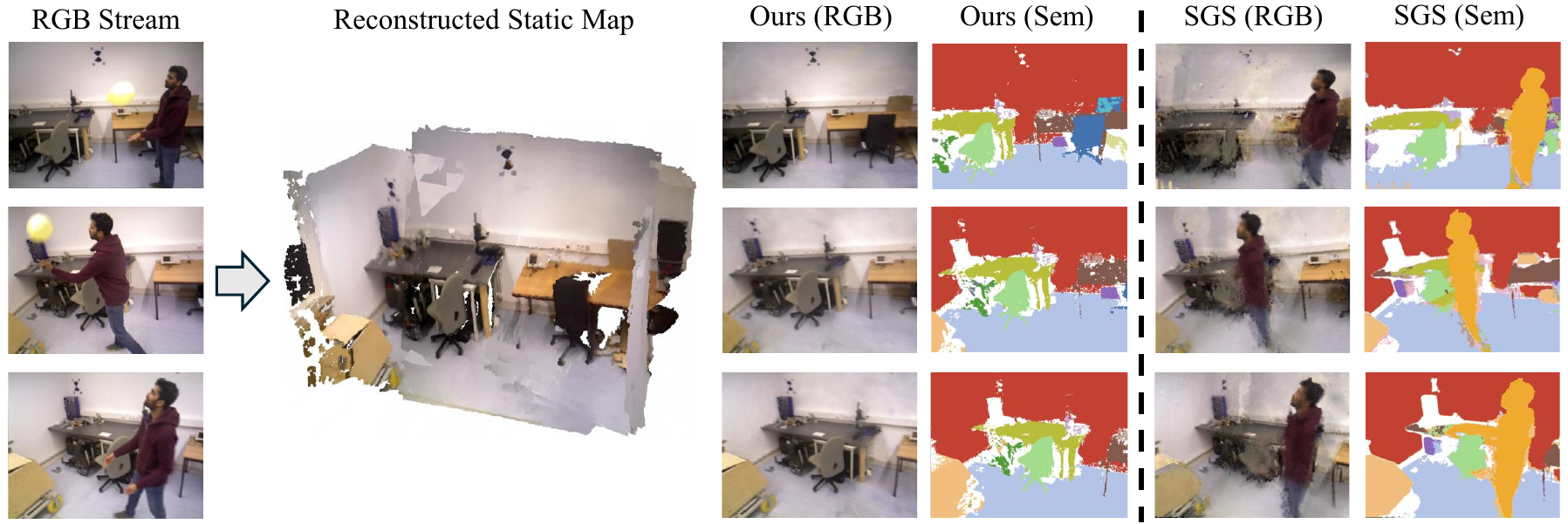}
	\caption{\textbf{DL-SLAM}. Given a monocular video sequence captured in dynamic environments, our method estimates accurate camera poses and reconstructs an artifact-free static map by leveraging dual-level dynamic probability to identify and remove dynamic objects. In contrast to SGS-SLAM, our method effectively eliminates the interference of dynamic objects, achieving high-fidelity rendering in both RGB and semantic views.}
	\label{fig1}
\end{teaserfigure}


\maketitle

\section{Introduction}
Visual SLAM is a foundational technology for applications such as embodied AI, autonomous driving, and mixed reality. Recent SLAM methods that employ Neural Radiance Fields (NeRF)~\cite{nerf} and 3D Gaussian Splatting (3DGS)~\cite{3dgs} as scene representations have demonstrated unprecedented capabilities in novel view synthesis and high-fidelity dense mapping~\cite{nice, splatam, gsicp, polar}. However, these approaches rely on a static world assumption, which causes their camera tracking and reconstruction accuracy to degrade significantly in the presence of dynamic objects. Dynamic objects introduce erroneous inter-frame correspondences that corrupt pose estimation, while their inclusion in the 3D map produces visual artifacts. 

A common strategy for mitigating this issue involves using semantic masks to uniformly exclude all objects within predefined dynamic categories, such as persons and vehicles~\cite{nid, rodyn, dg, sdd}. However, dynamic objects do not invariably exert detrimental effects on SLAM performance. When transiently static, they provide valuable geometric and photometric constraints for pose estimation. The indiscriminate removal of all masked regions leads to the unnecessary loss of potentially useful information. This information loss becomes critical when dynamic objects dominate the scene, as the remaining static structure may be too sparse to reliably constrain the camera pose. Moreover, reliance on a fixed set of predefined classes limits the generalizability of such methods to environments with dynamic objects outside the predefined categories.

Seeking to leverage the valuable information from transiently static objects, WildGS-SLAM~\cite{wildgs} employs per-pixel uncertainty maps to serve as weights in the tracking process, enabling transiently static objects to contribute to pose estimation. However, this approach exhibits two primary drawbacks. First, it fuses transiently static objects into the map representation due to their low uncertainty values, resulting in persistent artifacts. Second, the uncertainty estimation relies purely on geometric cues, leading to imprecise values at object boundaries, hindering the robust identification of dynamic regions.

To address these challenges, we present DL-SLAM, a monocular Gaussian Splatting SLAM system built upon a \textbf{D}ual-\textbf{L}evel probabilistic framework. Our method operates at two distinct scales. At the pixel level, an initial dynamic probability map is computed by combining fine-grained semantic labels with optical flow. This map serves as an adaptive weight within the dense bundle adjustment (DBA) layer, allowing stable regions of transiently static objects to contribute positively to tracking. At the object level, the pixel-wise probabilities are lifted to the 3D representation and aggregated according to semantic labels to calculate a holistic probability for each object. This object-level probability facilitates categorical pruning of dynamic Gaussians, yielding a coherent static map. Critically, these two levels are tightly coupled in a feedback loop. The purified static map is leveraged to render a geometrically consistent probability map, which in turn guides the Bayesian update of the initial pixel-level estimations. Furthermore, to achieve high-fidelity semantic mapping, we introduce a dynamic-aware semantic label refinement strategy, which handles occlusions caused by dynamic objects and rectifies label inconsistencies across frames. Extensive evaluations on three dynamic datasets show that DL-SLAM achieves state-of-the-art tracking accuracy and high-quality semantic mapping, surpassing existing dense dynamic SLAM systems.

Our main contributions are summarized as follows:
\begin{itemize}
\item A dual-level probabilistic framework for dynamic SLAM that enables transiently static objects to contribute to pose estimation while preventing their incorporation into the final map, thereby achieving robust tracking and artifact-free rendering in dynamic environments.
\item A feedback loop that tightly couples pixel-level and object-level processing, where dynamic Gaussian pruning at the object level guides the Bayesian update of per-pixel probabilities using the refined static map.
\item A dynamic-aware semantic label refinement strategy that compensates for dynamic object occlusion and resolves inter-frame semantic inconsistency, achieving high-fidelity semantic mapping.
\end{itemize}

\section{RELATED WORK} 
\subsection{Traditional Dynamic SLAM}
Early strategies for dynamic SLAM focus on excluding moving objects identified by semantic segmentation~\cite{ds, dyna, sad, blitz}. While these methods stabilized pose estimation by treating segmented regions as outliers, their classification of objects as either static or dynamic results in discarding valuable information from transiently static objects. To address this limitation, subsequent works introduce probabilistic frameworks. CFP-SLAM~\cite{cfp}, for instance, assigns each feature point a continuous static probability, updated over time using geometric constraints. This formulation offers greater robustness to segmentation errors and retains transiently static features, yet it remains dependent on predefined object categories and its multi-stage update can be computationally expensive. In contrast, our method leverages the 3DGS representation to integrate the dynamic probability directly as a Gaussian attribute. This allows for the identification of dynamic objects based on object-level probability rather than a predefined semantic categories, and enables efficient updates through the differentiable rendering pipeline.

\subsection{Dense Dynamic SLAM}
Recent works in dense dynamic SLAM have primarily focused on explicit motion filtering. Pioneering NeRF-based dynamic SLAM methods concentrate on robust motion masking, with NID-SLAM~\cite{nid} refining masks via depth and RoDyn-SLAM~\cite{rodyn} improving precision by fusing them with optical flow. This trend continues in 3DGS-based approaches. DG-SLAM~\cite{dg} introduces adaptive management of the Gaussians to handle dynamic objects, while SDD-SLAM~\cite{sdd} manages both active and passive dynamic objects. Nevertheless, the above approaches share two main limitations: 1) the dependence on prior semantic classes, which restricts generalization, and 2) the indiscriminate removal of potentially valuable constraints from transiently static objects. A notable exception is WildGS-SLAM~\cite{wildgs}, which uses learned uncertainty to leverage transiently static objects for pose estimation. However, it integrates transiently static objects that exhibit low uncertainty into the static map, and the uncertainty estimation can be imprecise at object boundaries. Our approach, in contrast, introduces dual-level probability to overcome these limitations. By incorporating semantic information, we estimate accurate pixel-level probability maps to leverage transiently static objects for robust tracking, while performing object-level pruning of dynamic objects to generate an artifact-free static map.

\begin{figure*}[t]
\centering
\includegraphics[width=0.98\linewidth]{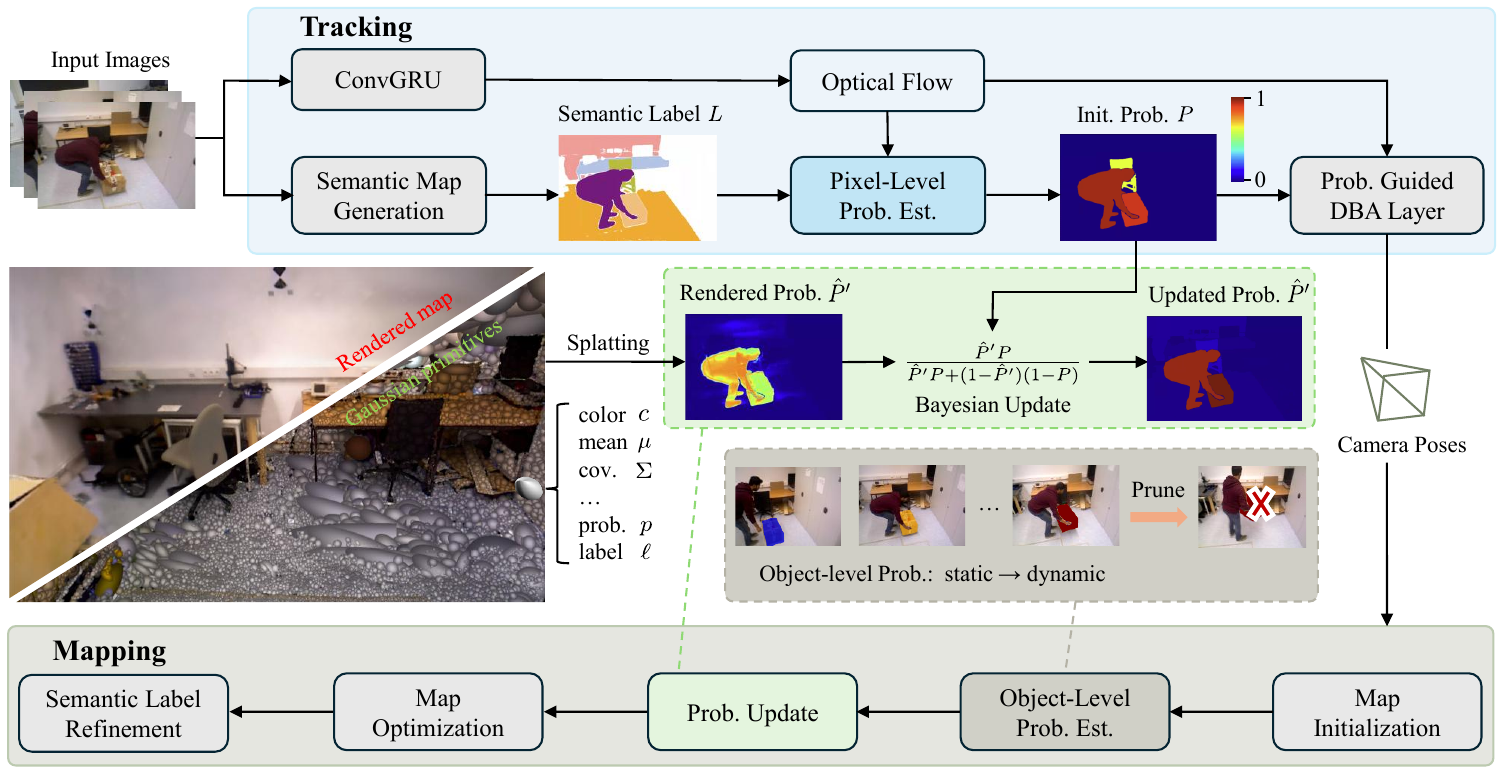}
\caption{\textbf{System Overview.} DL-SLAM takes RGB images as input to estimate camera poses and reconstruct a static 3D map. Our method leverages dual-level probability to mitigate interference from dynamic objects. The dynamic probability is a continuous score from 0 to 1, where higher values correspond to a greater likelihood of motion. In the tracking process, we use the semantic label map $L$ and optical flow to estimate a pixel-level dynamic probability map $P$, which serves as the weight of DBA optimization. In the mapping process, these probabilities and labels are lifted to the 3D Gaussian representation. This enables object-level probability estimation for dynamic Gaussian pruning. The static map is then used to render a probability map $\hat{P}'$ that, in turn, refines the initial estimate $P$.}
\label{fig2}
\end{figure*}

\subsection{Dense Semantic SLAM}
The integration of semantic information into 3DGS-based SLAM has advanced through three stages. SGS-SLAM~\cite{sgs}, as a pioneering effort, assigns additional semantic color parameters to each Gaussian, providing a straightforward yet coarse-grained semantic representation. Subsequent methods~\cite{gs3, semgauss, neds} pursued richer representations by embedding high-dimensional, implicit feature vectors within each Gaussian. While enabling differentiable optimization, these approaches necessitate a separate decoding stage to recover discrete labels, hindering direct object-level reasoning. A more recent paradigm, pioneered by OpenGS-SLAM~\cite{opengs}, advocates for an explicit representation by assigning a discrete class label to each Gaussian, facilitating object-level understanding. However, these methods are designed for static scenes. In dynamic environments, object motion and severe occlusions often result in temporal inconsistencies in the extracted semantic information, corrupting the semantic integrity of the map. To address this issue, we propose a dynamic-aware semantic refinement strategy that enforces temporal consistency by merging inconsistent and densifying occluded regions, ensuring high-fidelity semantic mapping.

\section{METHOD}

\subsection{Gaussian Scene Representation}
We utilize a 3D Gaussian representation \cite{3dgs} to reconstruct the static part of the scanned environment. The scene is represented by a set of anisotropic Gaussians $\mathcal{G}=\{g_i\}_{i=1}^K$. Each Gaussian $g_i$ is parameterized by its mean $\mu_i \in \mathbb{R}^3$, covariance $\Sigma_i \in \mathbb{R}^{3\times3}$, opacity $o_i \in \left[0,1\right]$, RGB color $c_i \in \mathbb{R}^3$. We additionally add dynamic probability $p_i \in \left[0,1\right]$ and explicit semantic label $\ell_i \in \mathbb{R}^1$ as Gaussian attributes. By using differentiable splatting to render color and depth maps, the scene is optimized via an iterative process that computes losses against input images and updates the Gaussian parameters accordingly.

The color image $\hat{C}$ and depth map $\hat{D}$ are rendered via alpha-blending proposed in~\cite{3dgs}:
\begin{equation}
\hat{C}=\sum_{i=1}^{n} c_i\alpha_iT_i ,\hat{D}=\sum_{i=1}^{n} d_i\alpha_iT_i,T_i=\prod_{j=1}^{i-1} \left(1-\alpha_j\right) ,
\label{eq1}
\end{equation}
where $c_i$ and $d_i$ denote the color and depth values of Gaussian $g_i$. Opacity $\alpha_i$ is computed by the coordinate $u$, mean $\mu$ and covariance $\Sigma_{2D}$ of the splatted 2D Gaussian in pixel space:
\begin{equation}
\alpha_i=o_i \exp \left(-\frac{1}{2}\left(u-\mu\right)^T \Sigma_{2D}^{-1}\left(u-\mu\right)\right).
\label{eq2}
\end{equation}

Similarly, the dynamic probability map is rendered as:
\begin{equation}
\hat{P}=\sum_{i=1}^{n} p_i\alpha_iT_i
\label{eq3}
\end{equation}

We render semantic label map $\hat{L}$ using Gaussian voting proposed in \cite{opengs}, where the label with the highest cumulative weight $W$ is assigned to the corresponding pixel $u$. The weight $W_j$ for label $l_j$ is computed as:
\begin{equation}
W_j=\sum_{g_i \in \mathcal{G}_u^j}\alpha_i\prod_{k=1}^{i-1} \left(1-\alpha_k\right), 
\label{eq4}
\end{equation}
where $\mathcal{G}_j^u$ is the set of Gaussians contributing to $u$ with $l_j$.

\subsection{Dual-Level Probabilistic Framework}
We propose a dual-level probabilistic framework to leverage the potential of transiently static objects, which couples 2D per-pixel estimation with 3D object-level reasoning.

\noindent\textbf{Semantic Map Generation.} The foundation of our framework is a robust and temporally consistent semantic understanding of the scene. Semantic label maps provide strong priors that ensure well-defined object boundaries in the pixel-level probability maps. More importantly, they form the basis for achieving object-level scene understanding.

Unlike methods that rely on predefined dynamic categories such as persons~\cite{rodyn, dg, sdd}, we employ the Recognize Anything Model~\cite{ram} to generate open-set class tags from the input image. These tags serve as text prompts for Grounding DINO~\cite{groundingdino} to generate class-aware bounding boxes, which then guide MobileSAMv2~\cite{mobilesamv2} to produce fine-grained semantic label maps. However, these per-frame segmentations lack temporal consistency across frames. To resolve this, we implement a robust data association method. Prior work~\cite{opengs} relies solely on Intersection over Union (IoU) for matching, which is often unreliable for dynamic objects with large displacements. We therefore compute a fused matching score combining both spatial overlap (IoU) and appearance similarity, using features extracted by CLIP~\cite{clip}. For each new detection, this fused score is computed against all existing tracked objects of the same semantic class. A match is established with the object yielding the highest score above threshold $\tau_\text{sim}$, inheriting its object ID. Detections that fail to match are initialized as new labels. The result is a semantic label map $L$, where each object is consistently identified across frames.

\noindent\textbf{Pixel-Level Probability Estimation.} To obtain a robust, pixel-wise dynamic probability map, we leverage epipolar geometry to quantify the degree of inconsistency for each pixel under the assumption of a static scene. For computational efficiency, the dynamic probability map is only estimated for keyframes. We employ a Convolutional Gated Recurrent Unit (ConvGRU)~\cite{droid} to estimate the optical flow between the current keyframe $I_k$ and the previous keyframe $I_{k-1}$, which is subsequently used for pose estimation. From the dense pixel correspondences derived from the optical flow, we estimate a fundamental matrix $F_{k,k-1}$. We then calculate the Sampson error $e^{F_{k,k-1}}(u)$, for each pixel $u$ to quantify its deviation from the epipolar constraint.

Inspired by CFP-SLAM \cite{cfp}, we model this geometric error statistically. Assuming the error for a static pixel follows a Gaussian distribution, the squared Sampson error can be modeled by a chi-squared distribution with two degrees of freedom. The dynamic probability for pixel $u$ is thus given by the cumulative distribution function (CDF) as:

\begin{equation}
\tilde{P}_{k|k-1}(u) = \text{CDF}{\chi^2}\left(\left(e^{F_{k,k-1}}(u)\right)^2; 2\right)
\label{eq5}
\end{equation}

This formulation yields an initial probability map $\tilde{P}_{k|k-1}$, where a larger geometric error corresponds to a higher likelihood of being dynamic. Under the assumption that all pixels on a rigid object share the same motion state, we perform a semantic aggregation for each object $o$. Let $p_o^\text{2D}$ denote the aggregated dynamic probability for object $o$ and $\mathcal{M}_o$ be its corresponding semantic mask. We collect the set of all per-pixel probabilities $\{\tilde{P}_{k|k-1}(u) | u \in \mathcal{M}_o\}$ and sort them in descending order. To ensure that significant motion dictates the overall object state, $p_o^\text{2D}$ is calculated as the average of the first quartile and the median of the sorted values. Finally, the aggregated dynamic probability map $P_k$ is constructed by assigning the probability $p_o^\text{2D}$ to all pixels within $\mathcal{M}_o$.
%, while unlabeled pixels are set to zero.

\noindent\textbf{Object-Level Probability Estimation.} By lifting the pixel-level information to the 3D representation as Gaussian attributes, our method enables object-level reasoning. Each new Gaussian $g_i$ , created from keyframe $k$, is associated with a dynamic probability $p_i$ and a semantic label $\ell_i$ from its corresponding pixel in $P_k$ and $L_k$.

A straightforward approach to compute the object-level probability would be averaging the dynamic probabilities of all associated Gaussians with an object. This method, however, suffers from temporal inertia, failing to handle situations where a long-term static object starts moving, as newly-added, high-probability Gaussians are diluted by a large history of static ones. To address this, we introduce a recency-weighted aggregation strategy. Let $\mathcal{G}_o$ denote the set of Gaussians associated with object $o$. At the current keyframe $k$, we assign an exponentially decaying weight $w_i=\exp(-\delta(k - k_i))$ to each Gaussian $g_i \in \mathcal{G}_o$, where $\delta=0.3$ is the decay rate and $k_i$ is the creation keyframe index of $g_i$. The object-level probability is then computed as:
\begin{equation}
p_o^\text{3D} = \frac{\sum_{g_i \in \mathcal{G}_o} w_i p_i}{\sum_{g_i \in \mathcal{G}_o} w_i}.
\label{eq6}
\end{equation}
This formulation ensures that the motion state of an object is predominantly dictated by its recent behavior, allowing for rapid detection of state changes.

To maintain a coherent and static scene representation, we remove dynamic objects from the Gaussian map. To avoid premature removal of static objects, we employ a temporal confirmation protocol. An object is marked as dynamic and added to a global dynamic object set if $p_o^\text{3D}$ exceeds a threshold $\tau_\text{prune}$ for three consecutive keyframes. All Gaussians belonging to any object within this global dynamic set are removed from the scene.

\noindent\textbf{Probability Update.} The fidelity of the aggregated probability map $P_k$ is inherently limited by the reliability of the initial estimation $\tilde{P}_{k|k-1}$. Since these initial estimates are derived from 2D optical flow, they are prone to error in textureless regions with poor feature quality. Furthermore, traditional geometry-based methods~\cite{cfp} rely on costly multi-stage pipelines to update the initial probability.

To address these limitations, we introduce an efficient update method that leverages the 3D Gaussian representation. The core of this approach is a feedback loop that tightly couples the pixel and object levels of our framework, using the optimized 3D map to correct the frame-to-frame 2D estimates. Specifically, we render a geometrically consistent likelihood map $\hat{P}_k$ using Eq.~\ref{eq3} from the 3D representation. However, the rendering process acts as a weighted average, which diminishes the contribution of high-probability dynamic Gaussians as they are blended with numerous static ones. To amplify the influence of high-probability values, we apply a power-law scaling to the rendered map as $\hat{P}_k' = (\hat{P}_k)^\gamma$, where $\gamma$ is set to 0.5. The amplified probability $\hat{P}'_k$ is then fused with the initial estimate $P_k$ within a Bayesian framework to compute a posterior probability as:
\begin{equation}
P_k^\text{post} = \frac{\hat{P}'_k \cdot P_k}{\hat{P}'_k \cdot P_k + (1-\hat{P}'_k) \cdot (1-P_k)}.
\label{eq7}
\end{equation}

To maintain consistency, the 2D posterior probability $P_k^\text{post}$ is propagated back to the 3D representation. For the current viewpoint, the 3D centers of all visible Gaussians are projected onto the image plane. The corresponding pixel-wise probabilities are used to update the dynamic attribute $p_i$ of each Gaussian $g_i$.

\subsection{Semantic Label Refinement}
\begin{figure}[t]
\begin{minipage}[c]{0.32\linewidth}		
	\centering\small\text{Input Label Map}
\end{minipage}
\begin{minipage}[c]{0.32\linewidth}		
	\centering\small\text{w/o Refinement}
\end{minipage}
\begin{minipage}[c]{0.32\linewidth}		
	\centering\small\text{w/ Refinement}
\end{minipage}

\begin{minipage}[c]{0.32\linewidth}
	\includegraphics[width=0.98\linewidth]{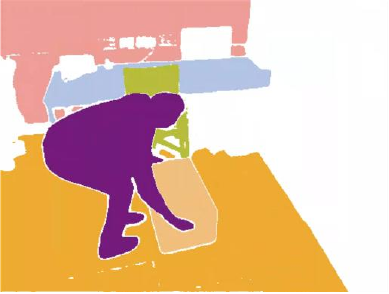}
\end{minipage}
\begin{minipage}[c]{0.32\linewidth}
	\includegraphics[width=0.98\linewidth]{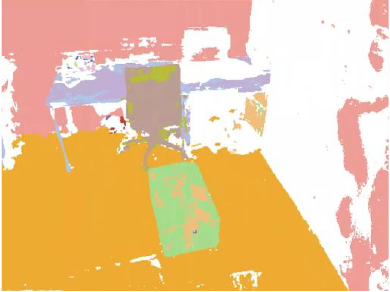}
\end{minipage}
\begin{minipage}[c]{0.32\linewidth}
	\includegraphics[width=0.98\linewidth]{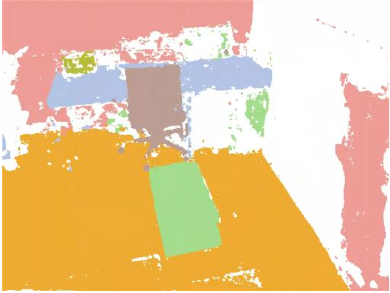}
\end{minipage}

\caption{\textbf{Effect of Semantic Label Refinement.} Our method resolves semantic inconsistencies (e.g., the box on the floor) and robustly reconstructs regions occluded by dynamic objects (e.g., the chair).}
\label{fig3}
\end{figure}

While integrating semantic labels into 3D reconstruction enables object-level scene understanding, dynamic environments pose two primary challenges. First, despite our 2D data association, lingering semantic inconsistencies can still arise, as the matching relies purely on frame-to-frame cues and lacks the global geometric context provided by the 3D map. Second, dynamic objects cause severe occlusions that lead to sparse and under-reconstructed regions, characterized by semantically ambiguous Gaussians. As shown in Fig.~\ref{fig3}, our method effectively corrects input labels and plausibly completes the regions occluded by dynamic objects.

To resolve semantic inconsistencies, we introduce a robust online correction mechanism. The goal is to determine if a rendered object mask with label $i$ and an input mask with label $j$ correspond to the same object. To address label ambiguities, we match each rendered mask $\mathcal{M}_i \subset \hat{L}$ with input mask $\mathcal{M}_j \subset L$ using a score function that combines IoU with containment ratios:
\begin{equation}
S_\text{m}(i, j) = \max\left(\frac{|\mathcal{M}_i \cap \mathcal{M}_j|}{|\mathcal{M}_i \cup \mathcal{M}_j|}, \frac{|\mathcal{M}_i \cap \mathcal{M}_j|}{|\mathcal{M}_j|}, \frac{|\mathcal{M}_i \cap \mathcal{M}_j|}{|\mathcal{M}_i|}\right).
\end{equation}
If the matching score $S_\text{m}(i, j)$ exceeds a threshold $\tau_\text{match}$, we establish a potential match, considering label $j$ to be an alias for label $i$. To ensure robustness against transient errors, this association is only solidified into a permanent identity merge after the same label pair matches for three consecutive keyframes. Once merged, the Gaussians are then relabeled to enforce object-level consistency.

To tackle the challenge of under-reconstructed regions, we introduce a dynamic-aware densification strategy. We target regions exhibiting high semantic fragmentation that lie within the aggregation of all dynamic object masks. Semantic fragmentation is quantified using a pixel-wise semantic gradient $S_\text{g}(u)$, which counts the number of neighbors in a 4-connected neighborhood $\mathcal{N}(u)$ with a different semantic label as:
\begin{equation}
S_\text{g}(u) = \sum_{q \in \mathcal{N}(u)} \mathbb{I}(\hat{L}(u) \neq \hat{L}(q)),
\end{equation}
where $\mathbb{I}(\cdot)$ is the indicator function. Densification is performed on regions where $S_\text{g}(u)$ exceeds the threshold $\tau_\text{grad}$, thereby improving the reconstruction quality of regions corrupted by dynamic objects.

\subsection{SLAM process}
\textbf{Tracking.} Our tracking module builds upon DROID-SLAM~\cite{droid}, which employs a DBA layer to jointly optimize keyframe poses $\omega$ and disparities $d$ over a frame graph $F = (V, E)$. Here, $V$ represents the set of keyframes and $E$ represents the keyframe edges. To enhance robustness against dynamic objects, we incorporate our dynamic probability map as a weighting term in the DBA formulation. We estimate the metric depth $\tilde{D}_i$ using Metric3D~\cite{metric3dv2} for each new keyframe $I_i$ and incorporate it into the DBA objective:

\begin{equation}
%\begin{split}
\operatorname*{arg\,min}_{\omega,\,d}\sum_{(i,j)\in E}
%&
\Bigl\lVert
\tilde{\mathbf{u}}_{ij}-\Pi_c\!\bigl(\omega_j^{-1}\,\omega_i\,\Pi_c^{-1}(u_i,d_i)\bigr)
\Bigr\rVert_{\Sigma_{ij}/(1-P_i)}^{2}.
%\\&+
%\lambda_{4}\,\sum_{i\in V}\Bigl\lVert M_i\bigl(d_i-1/\tilde{D}_i\bigr)\Bigr\rVert^{2},
%\end{split}
\label{eq8}
\end{equation}
where $u_i$ is the pixel coordinates of keyframe $I_i$ and $\tilde{\mathbf{u}}_{ij}$ is the predicted pixel correspondence with keyframe $I_j$ by optical flow estimation. $\Pi_c$ is the 3D-2D camera projection function. The error terms are weighted by the confidence matrix $\Sigma_{ij}$ from ConvGRU and our dynamic probability map $P_i$. This formulation ensures that pixels exhibiting active motion are down-weighted, while stable regions of transiently static objects can contribute to the DBA.

\noindent\textbf{Mapping.} We incrementally grow the Gaussian map $\mathcal{G}$ with new keyframes to cover newly observed regions. Unlike previous methods that attempt to filter dynamic objects in 2D~\cite{slamx, sdd}, our approach allows all pixels to initialize new Gaussians and handle the pruning of dynamic Gaussians in the 3D representation.

Following map expansion, the Gaussian map is optimized over a local window of covisible keyframes. For each selected keyframe, we render the color $\hat{C}$ and depth $\hat{D}$ images from the Gaussian model and minimize the rendering loss:

\begin{equation}
\left(1-P\right) \left(\lambda_\text{c} \mathcal{L}_{\text{color}} + \left(1-\lambda_\text{c}\right)\mathcal{L}_{\text{depth}}\right) + \lambda_\text{reg} \mathcal{L}_{\text{iso}}.
\label{eq9}
\end{equation}
The color loss $\mathcal{L}_{\text{color}}$ combines L1 and SSIM losses as:
\begin{equation}
\mathcal{L}_{\text{color}}=\left(1-\lambda_\text{ssim}\right)|\hat{C}-C|_1+\lambda_\text{ssim}\left(1-\operatorname{SSIM}(\hat{C}, C)\right) \text {. }
\label{eq10}
\end{equation}
The depth loss is defined as: 
\begin{equation}
\mathcal{L}_{\text{depth}}=|\hat{D}-\tilde{D}|_1.
\label{eq:11}
\end{equation}
To mitigate the influence of dynamic objects, both the color loss and the depth loss terms are weighted by our dynamic probability map $P$. The isotropic loss $\mathcal{L}_{\text{iso}}$~\cite{monogs} is included to prevent Gaussians in less constrained regions from becoming excessively elongated.
\begin{equation}
\mathcal{L}_{\text{reg}}=\frac{1}{|\mathcal{G}|}\sum_{i=1}^{|\mathcal{G}|}|s_i-\bar{s}_i|_1,
\label{eq:12}
\end{equation}
where $\mathcal{G}$ denotes the Gaussian map, $s_i$ is the scale of the $i$-th Gaussian, $\bar{s}_i$ represents the mean scale within the map. Instead of optimizing for a fixed number of iterations, we terminate the optimization once the variance of the mapping loss in the local window falls below $\tau_\text{var}$, avoiding redundant iterations after convergence.

\begin{figure*}[ht]
\centering
\begin{minipage}[c]{0.04\linewidth}
	\centering
	\rotatebox{90}{}
\end{minipage}
\begin{minipage}[c]{0.13\linewidth}
	\centering\small\text{Ground Truth}
\end{minipage}
\begin{minipage}[c]{0.13\linewidth}
	\centering\small\text{SGS-SLAM}
\end{minipage}
\begin{minipage}[c]{0.13\linewidth}
	\centering\small\text{DG-SLAM}
\end{minipage}
\begin{minipage}[c]{0.26\linewidth}
	\centering\small\text{WildGS-SLAM}
\end{minipage}
\begin{minipage}[c]{0.26\linewidth}
	\centering\small\textbf{DL-SLAM (Ours)}
\end{minipage}

\centering
\begin{minipage}[c]{0.04\linewidth}
	\centering
	\rotatebox{90}{\small\texttt{wandering}}
\end{minipage}
\begin{minipage}[c]{0.13\linewidth}
	\centering
	\includegraphics[width=0.98\linewidth]{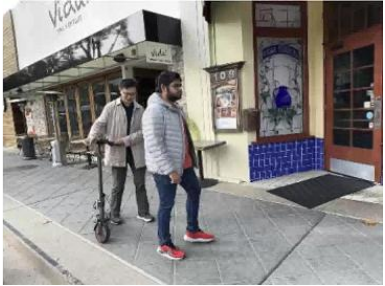}
\end{minipage}
\begin{minipage}[c]{0.13\linewidth}
	\centering
	\includegraphics[width=0.98\linewidth]{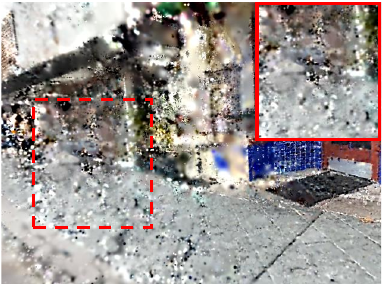}
\end{minipage}
\begin{minipage}[c]{0.13\linewidth}
	\centering
	\includegraphics[width=0.98\linewidth]{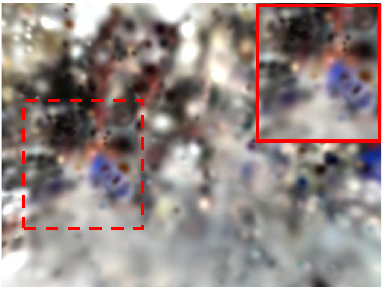}
\end{minipage}
\begin{minipage}[c]{0.13\linewidth}
	\centering
	\includegraphics[width=0.98\linewidth]{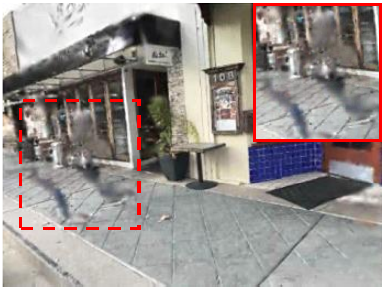}
\end{minipage}
\begin{minipage}[c]{0.13\linewidth}
	\centering
	\includegraphics[width=0.98\linewidth]{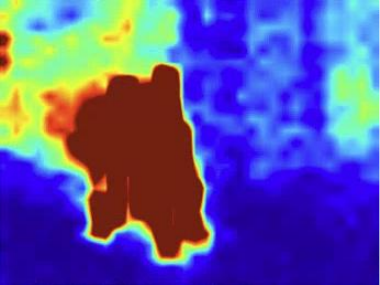}
\end{minipage}
\begin{minipage}[c]{0.13\linewidth}
	\centering
	\includegraphics[width=0.98\linewidth]{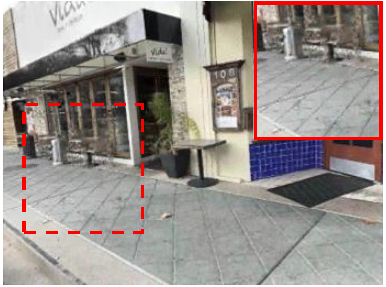}
\end{minipage}
\begin{minipage}[c]{0.13\linewidth}
	\centering
	\includegraphics[width=0.98\linewidth]{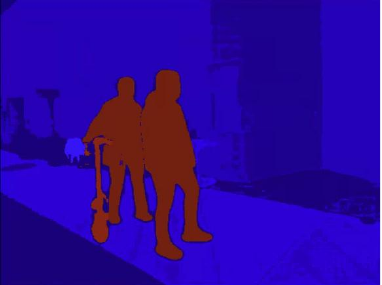}
\end{minipage}

\centering
\begin{minipage}[c]{0.04\linewidth}
	\centering
	\rotatebox{90}{\small\texttt{street}}
\end{minipage}
\begin{minipage}[c]{0.13\linewidth}
	\centering
	\includegraphics[width=0.98\linewidth]{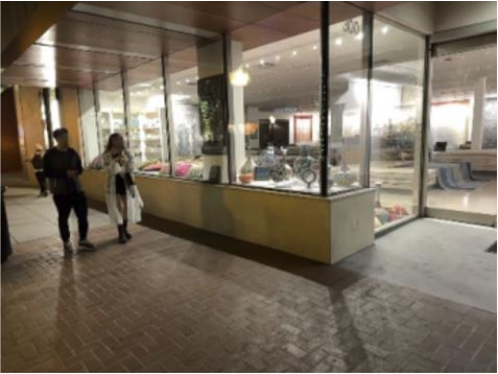}
\end{minipage}
\begin{minipage}[c]{0.13\linewidth}
	\centering
	\includegraphics[width=0.98\linewidth]{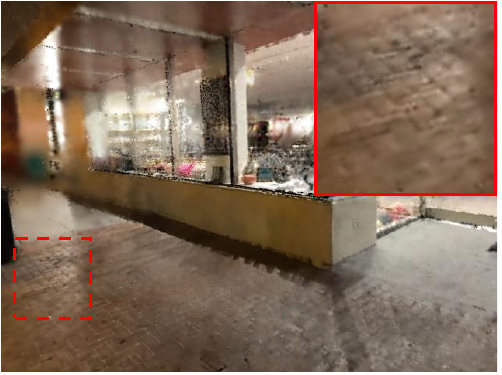}
\end{minipage}
\begin{minipage}[c]{0.13\linewidth}
	\centering
	\includegraphics[width=0.98\linewidth]{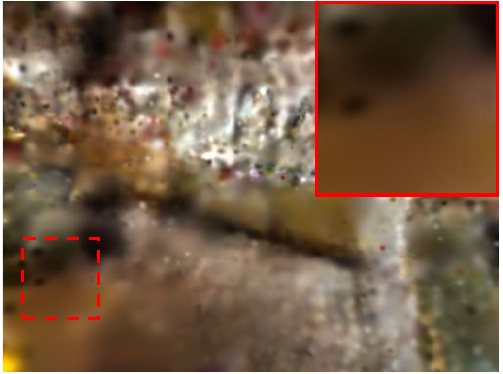}
\end{minipage}
\begin{minipage}[c]{0.13\linewidth}
	\centering
	\includegraphics[width=0.98\linewidth]{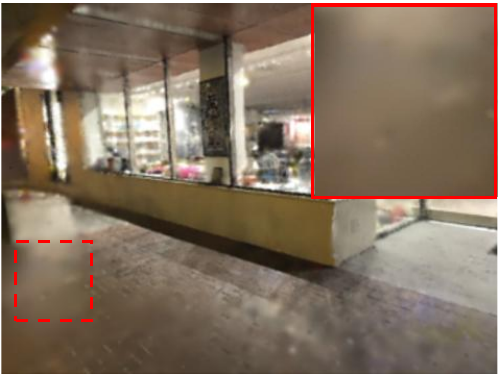}
\end{minipage}
\begin{minipage}[c]{0.13\linewidth}
	\centering
	\includegraphics[width=0.98\linewidth]{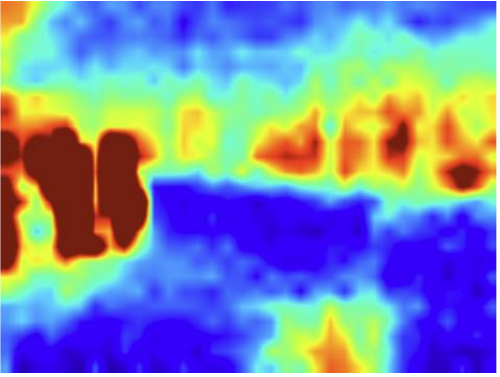}
\end{minipage}
\begin{minipage}[c]{0.13\linewidth}
	\centering
	\includegraphics[width=0.98\linewidth]{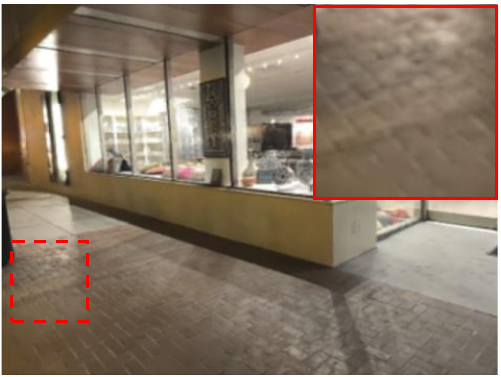}
\end{minipage}
\begin{minipage}[c]{0.13\linewidth}
	\centering
	\includegraphics[width=0.98\linewidth]{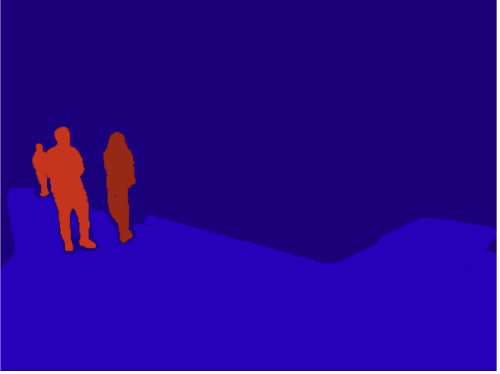}
\end{minipage}

\centering
\begin{minipage}[c]{0.04\linewidth}
	\centering
	\rotatebox{90}{\small\texttt{ps\_trk}}
\end{minipage}
\begin{minipage}[c]{0.13\linewidth}
	\centering
	\includegraphics[width=0.98\linewidth]{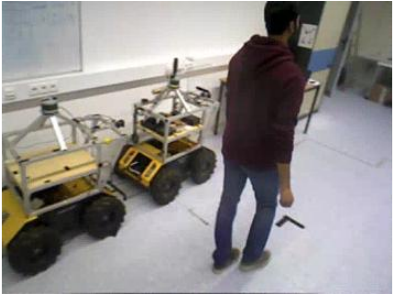}
\end{minipage}
\begin{minipage}[c]{0.13\linewidth}
	\centering
	\includegraphics[width=0.98\linewidth]{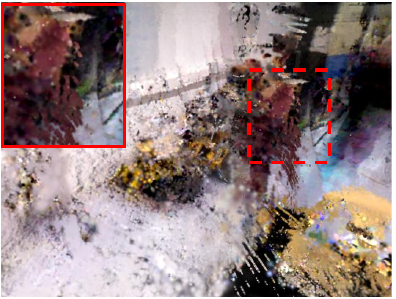}
\end{minipage}
\begin{minipage}[c]{0.13\linewidth}
	\centering
	\includegraphics[width=0.98\linewidth]{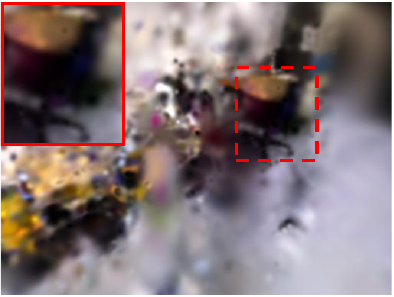}
\end{minipage}
\begin{minipage}[c]{0.13\linewidth}
	\centering
	\includegraphics[width=0.98\linewidth]{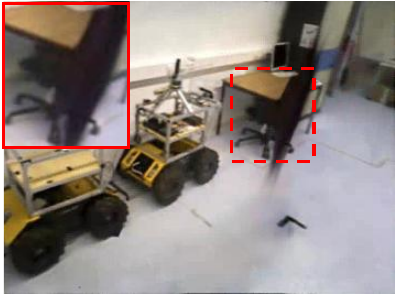}
\end{minipage}
\begin{minipage}[c]{0.13\linewidth}
	\centering
	\includegraphics[width=0.98\linewidth]{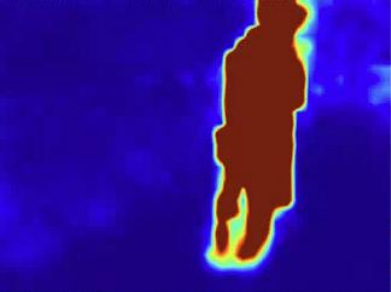}
\end{minipage}
\begin{minipage}[c]{0.13\linewidth}
	\centering
	\includegraphics[width=0.98\linewidth]{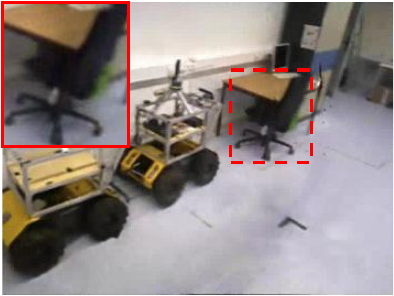}
\end{minipage}
\begin{minipage}[c]{0.13\linewidth}
	\centering
	\includegraphics[width=0.98\linewidth]{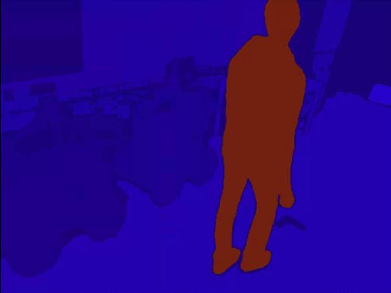}
\end{minipage}

\centering
\begin{minipage}[c]{0.04\linewidth}
	\centering
	\rotatebox{90}{\small\texttt{ps\_trk2}}
\end{minipage}
\begin{minipage}[c]{0.13\linewidth}
	\centering
	\includegraphics[width=0.98\linewidth]{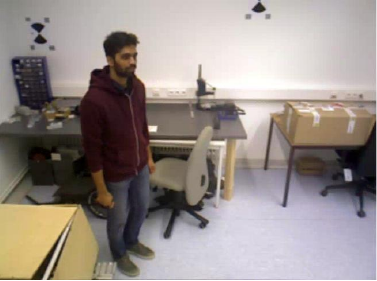}
\end{minipage}
\begin{minipage}[c]{0.13\linewidth}
	\centering
	\includegraphics[width=0.98\linewidth]{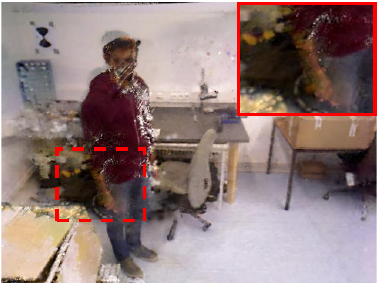}
\end{minipage}
\begin{minipage}[c]{0.13\linewidth}
	\centering
	\includegraphics[width=0.98\linewidth]{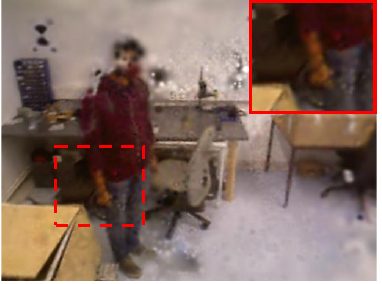}
\end{minipage}
\begin{minipage}[c]{0.13\linewidth}
	\centering
	\includegraphics[width=0.98\linewidth]{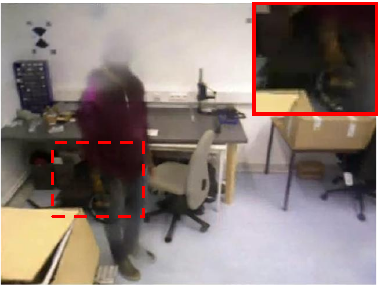}
\end{minipage}
\begin{minipage}[c]{0.13\linewidth}
	\centering
	\includegraphics[width=0.98\linewidth]{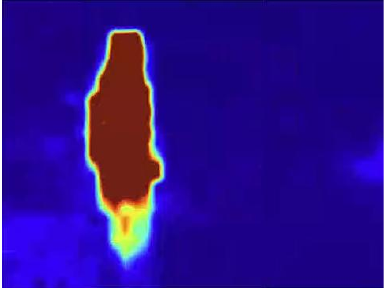}
\end{minipage}
\begin{minipage}[c]{0.13\linewidth}
	\centering
	\includegraphics[width=0.98\linewidth]{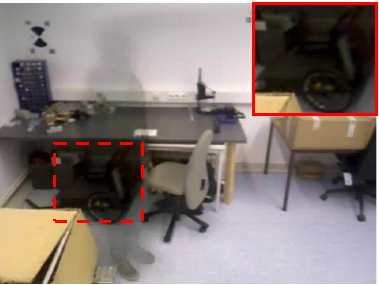}
\end{minipage}
\begin{minipage}[c]{0.13\linewidth}
	\centering
	\includegraphics[width=0.98\linewidth]{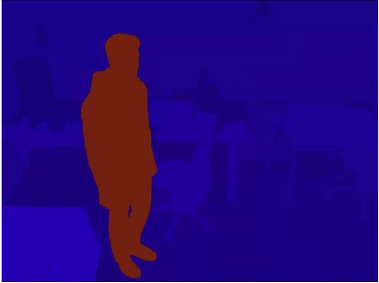}
\end{minipage}

\centering
\begin{minipage}[c]{0.04\linewidth}
	\centering
	\rotatebox{90}{\small\texttt{mv\_box}}
\end{minipage}
\begin{minipage}[c]{0.13\linewidth}
	\centering
	\includegraphics[width=0.98\linewidth]{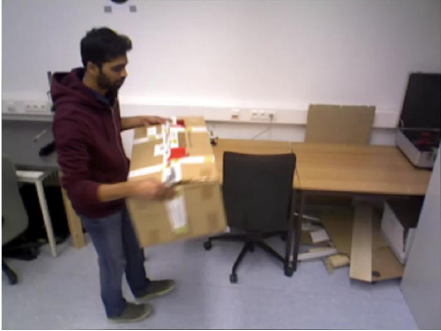}
\end{minipage}
\begin{minipage}[c]{0.13\linewidth}
	\centering
	\includegraphics[width=0.98\linewidth]{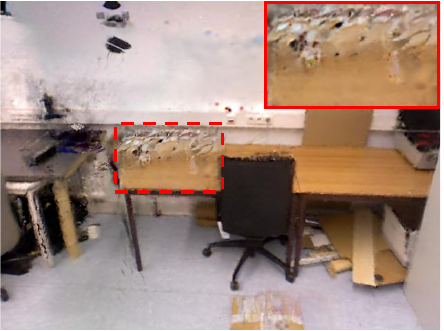}
\end{minipage}
\begin{minipage}[c]{0.13\linewidth}
	\centering
	\includegraphics[width=0.98\linewidth]{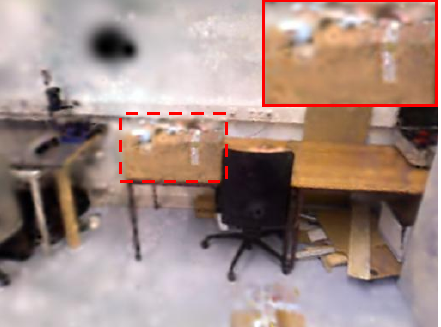}
\end{minipage}
\begin{minipage}[c]{0.13\linewidth}
	\centering
	\includegraphics[width=0.98\linewidth]{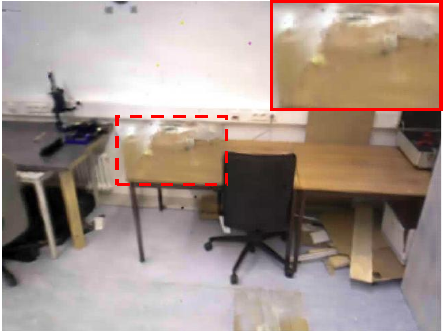}
\end{minipage}
\begin{minipage}[c]{0.13\linewidth}
	\centering
	\includegraphics[width=0.98\linewidth]{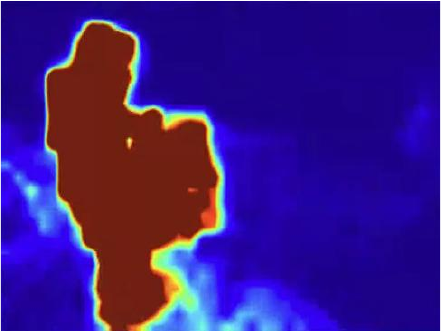}
\end{minipage}
\begin{minipage}[c]{0.13\linewidth}
	\centering
	\includegraphics[width=0.98\linewidth]{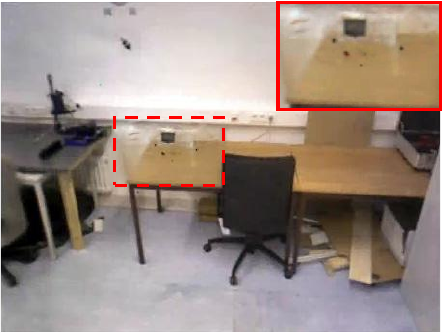}
\end{minipage}
\begin{minipage}[c]{0.13\linewidth}
	\centering
	\includegraphics[width=0.98\linewidth]{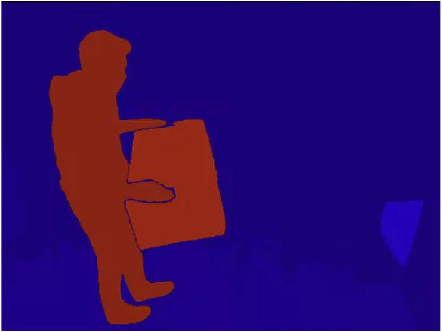}
\end{minipage}

\caption{\textbf{Rendering results on Wild-SLAM iPhone and BONN datasets}. Alongside the rendered images, we compare the uncertainty map of WildGS-SLAM against our dynamic probability map. The red box shows the zoomed-in view of the red dashed box. WildGS-SLAM suffers from artifacts due to transiently static objects (e.g., the two persons in $\texttt{wandering}$ and the box in $\texttt{mv\_box}$), whereas our method achieves artifact-free rendering via object-level pruning.}
\label{fig4}
\end{figure*}

\begin{table*}[t]
	\centering
	\caption{\textbf{Rendering performance on Wild-SLAM iPhone and BONN datasets}. The best results are indicated in \textbf{bold}.}\label{tab3}
	\begin{tabular}{llcccccccccc}
		\toprule
		\multirow{2}{*}{Method} & \multirow{2}{*}{Metric} & \multicolumn{6}{c}{Wild-SLAM iPhone} & \multicolumn{4}{c}{BONN}\\
		\cmidrule(lr){3-8}\cmidrule(lr){9-12}
		& & \texttt{piano} & \texttt{shop} & \texttt{street} & \texttt{tower} & \texttt{wall} & \texttt{wander} & \texttt{ps\_trk} & \texttt{ps\_trk2} & \texttt{mv\_box} & \texttt{mv\_box2}\\
		% \midrule
		% \multirow{4}*{Splat-SLAM} & PSNR[dB]$\uparrow$ &  &  &  &  & 17.08 &  &  &  & \\
		%  & SSIM[\%]$\uparrow$ &  &  &  &  & 0.652 &  &  &  & \\
		%  & LPIPS[\%]$\downarrow$ &  &  &  &  & 0.480 &  &  &  & \\
		%  & Depth L1[cm]$\downarrow$ &  &  &  &  & 29.33 &  &  &  & \\
		\midrule
		\multirow{3}*{SGS-SLAM} & PSNR[dB]$\uparrow$ & 16.65 & 15.59 & 18.31 & 17.53 & 14.15 & 14.11 & 17.02 & 15.78 & 20.46 & 19.68\\
		& SSIM[\%]$\uparrow$ & 0.669 & 0.584 & 0.736 & 0.715 & 0.415 & 0.513 & 0.635 & 0.604 & 0.829 & 0.785\\
		& LPIPS[\%]$\downarrow$ & 0.409 & 0.583 & 0.418 & 0.385 & 0.681 & 0.561 & 0.313 & 0.341 & \textbf{0.193} & 0.232\\
		\midrule
		\multirow{3}*{DG-SLAM} & PSNR[dB]$\uparrow$ & 13.41 & 12.76 & 13.23 & 13.98 & 11.97 & 13.14 & 17.11 & 16.38 & 20.92 & 21.11\\
		& SSIM[\%]$\uparrow$ & 0.263 & 0.227 & 0.375 & 0.531 & 0.203 & 0.260 & 0.666 & 0.686 & 0.776 & 0.798\\
		& LPIPS[\%]$\downarrow$ & 0.846 & 0.936 & 0.855 & 0.674 & 0.994 & 0.831 & 0.656 & 0.610 & 0.423 & 0.422\\
		\midrule
		\multirow{3}*{WildGS-SLAM} & PSNR[dB]$\uparrow$ & 17.93 & 15.79 & 19.69 & 19.06 & 16.32 & 16.70 & 20.86 & 20.46 & 22.55 & 21.45\\
		& SSIM[\%]$\uparrow$ & 0.650 & 0.499 & 0.700 & 0.805 & 0.503 & 0.569 & 0.854 & 0.853 & 0.866 & 0.800\\
		& LPIPS[\%]$\downarrow$ & 0.248 & 0.468 & 0.337 & 0.245 & 0.493 & 0.341 & 0.260 & 0.239 & 0.213 & 0.401\\
%		& Depth L1[cm]$\downarrow$ & 39.04 & 29.82 & 141.68 & 221.9 & 26.45 & 34.88 & 14.97 & 44.99 & 69.21\\
		\midrule
		\multirow{3}*{\textbf{DL-SLAM (Ours)}}& PSNR[dB]$\uparrow$ & \textbf{19.18} & \textbf{18.34} & \textbf{21.06} & \textbf{19.93} & \textbf{18.07} & \textbf{17.92} & \textbf{21.81} & \textbf{21.37} & \textbf{23.74} & \textbf{22.62}\\
		& SSIM[\%]$\uparrow$ & \textbf{0.764} & \textbf{0.689} & \textbf{0.810} & \textbf{0.864} & \textbf{0.715} & \textbf{0.751} & \textbf{0.859} & \textbf{0.859} & \textbf{0.869} & \textbf{0.862}\\
		& LPIPS[\%]$\downarrow$ & \textbf{0.171} & \textbf{0.236} & \textbf{0.175} & \textbf{0.163} & \textbf{0.236} & \textbf{0.197} & \textbf{0.253} & \textbf{0.235} & {0.208} & \textbf{0.229}\\
%		& Depth L1[cm]$\downarrow$ & \textbf{24.09} & \textbf{20.77} & \textbf{81.93} & \textbf{93.30} & \textbf{24.14} & \textbf{22.62} & \textbf{13.97} & \textbf{24.99} & \textbf{38.23}\\
		\bottomrule
	\end{tabular}
\end{table*}

\begin{table*}[t]
	 \setlength\tabcolsep{3pt}
	\centering
	\caption{\textbf{Tracking performance on BONN dataset} (ATE [cm]$\downarrow$). The best results are indicated in \textbf{bold}, and the second-best results are \underline{underlined}. ``-'' indicates the absence of mention.}\label{tab1}
	\begin{tabular}{lcccccccccccccccccc}
		\toprule
		\multirow{2}{*}{Method} & \multicolumn{2}{c}{\texttt{ball}} & \multicolumn{2}{c}{\texttt{ball2}} & \multicolumn{2}{c}{\texttt{crowd}} & \multicolumn{2}{c}{\texttt{crowd2}} & \multicolumn{2}{c}{\texttt{ps\_trk}} & \multicolumn{2}{c}{\texttt{ps\_trk2}} & \multicolumn{2}{c}{\texttt{mv\_box}} & \multicolumn{2}{c}{\texttt{mv\_box2}} & \multicolumn{2}{c}{Avg.}\\
		\cmidrule(lr){2-3}\cmidrule(lr){4-5}\cmidrule(lr){6-7}\cmidrule(lr){8-9}\cmidrule(lr){10-11}\cmidrule(lr){12-13}\cmidrule(lr){14-15}\cmidrule(lr){16-17}\cmidrule(lr){18-19}
		& RMSE & S.D. & RMSE & S.D. & RMSE & S.D. & RMSE & S.D. & RMSE & S.D. & RMSE & S.D. & RMSE & S.D. & RMSE & S.D. & RMSE & S.D.\\
		\midrule
		\rowcolor{gray!10}
		\multicolumn{19}{l}{\textit{Traditional}}\\
		ORB-SLAM2 & 15.8 & 6.3 & 12.6 & 5.4 & 98.0 & 60.1 & 62.8 & 35.7 & 69.8 & 32.7 & 78.6 & 45.4 & 32.3 & 11.5 & 79.6 & 35.9 & 56.2 & 29.1\\
		% DSO & RGB-D & 7.3 & 21.8 & 10.1 & 7.6 & 30.6 & 26.5 & 4.7 & 11.2 & 15.0\\
%		DROID-SLAM & 7.5 & - & 4.1 & - & 5.2 & - & 6.5 & - & 4.3 & - & 5.4 & - & 2.3 & - & 4.0 & - & 4.9 & -\\
		ReFusion & 17.5 & - & 25.4 & - & 20.4 & - & 15.5 & - & 28.9 & - & 46.3 & - & 7.1 & - & 17.9 & - & 22.4 & -\\
		DynaSLAM & 3.4 & 1.5 & 3.2 & 1.4 & 2.5 & 1.3 & 2.9 & 1.7 & 4.7 & \textbf{1.6} & 12.6 & 7.2 & 2.0 & 1.0 & 26.2 & 11.8 & 7.2 & 3.4\\
		\midrule
		\rowcolor{gray!10}
		\multicolumn{19}{l}{\textit{NeRF-based}}\\
%		NICE-SLAM & 24.4 & 20.2 & 19.3 & 35.8 & 24.5 & 53.6 & 17.7 & 8.3 & 22.7\\
		Co-SLAM & 25.0 & 12.4 & 38.7 & 20.8 & 245.8 & 64.5 & 229.0 & 103.6 & 77.2 & 35.6 & 73.7 & 31.1 & 10.5 & 4.6 & 25.4 & 11.5 & 90.7 & 35.5\\
		RoDyn-SLAM & 7.9 & 2.7 & 11.5 & 6.1 & 13.5 & 5.9 & 13.7 & 4.5 & 14.5 & 4.6 & 13.8 & 3.5 & 24.9 & 13.6 & 12.6 & 4.7 & 14.1 & 5.7\\
%		DDN-SLAM & \textbf{1.8} & 4.1 & 1.8 & 2.3 & 4.3 & 3.8 & 2.0 & 3.2 & 2.9\\
		DynaMoN & 2.9 & \underline{1.2} & 2.7 & \textbf{1.0} & 3.3 & 2.6 & 4.8 & 3.5 & 15.1 & 9.7 & \textbf{2.3} & \textbf{0.9} & \underline{1.5} & \textbf{0.7} & 2.7 & \textbf{1.0} & 4.4 & 2.6\\
		\midrule
		\rowcolor{gray!10}
		\multicolumn{19}{l}{\textit{3DGS-based}}\\
%		SplaTAM & 35.9 & 14.5 & 34.0 & 15.7 & 243.2 & 54.3 & 172.4 & 75.7 & 120.9 & 23.4 & 99.6 & 49.2 & 6.2 & 6.6 & 16.3 & 9.6 & - & -\\
		SGS-SLAM & 44.7 & 20.1 & 37.2 & 18.9 & 105.8 & 28.9 & 267.0 & 78.1 & 128.7 & 36.6 & 66.7 & 36.8 & 23.6 & 22.7 & 41.5 & 20.8 & 89.4 & 32.9\\
		% MonoGS & RGB-D & 15.3 & 17.3 & 11.3 & 7.3 & 26.4 & 35.2 & 22.2 & 47.2 & 22.8\\
%		Splat-SLAM & 8.8 & - & 3.0 & - & 6.8 & - & 7.7 & - & 4.9 & - & 25.8 & - & \underline{1.7} & - & 3.0 & - & 7.7 & -\\
		DG-SLAM & 6.2 & 2.3 & 4.8 & 2.0 & 2.1 & 1.8 & 3.2 & 2.5 & 4.6 & 3.4 & 6.3 & 2.6 & 2.8 & 1.9 & 3.6 & 1.8 & 4.2 & 2.3\\
%		PG-SLAM & 6.4 & 2.2 & 7.3 & 3.4 & - & - & - & - & 5.0 & 1.9 & 8.5 & 2.8 & 4.6 & 1.3 & 7.0 & 2.0 & - & -\\
		SDD-SLAM & \textbf{2.2} & - & 4.0 & - & - & - & - & - & 4.5 & - & 6.5 & - & - & - & \textbf{2.1} & - & - & -\\
		WildGS-SLAM & 2.8 & \underline{1.2} & \underline{2.5} & \underline{1.1} & \underline{1.6} & \underline{0.8} & \underline{2.2} & \underline{1.3} & \underline{3.7} & \underline{1.8} & {3.1} & 1.4 & {1.7} & \underline{0.9} & \underline{2.3} & \underline{1.2} & \underline{2.5} & \underline{1.2}\\
		\textbf{Ours} & \underline{2.4} & \textbf{1.1} & \textbf{2.3} & \textbf{1.0} & \textbf{1.3} & \textbf{0.6} & \textbf{1.6} & \textbf{0.7} & \textbf{3.4} & \textbf{1.6} & \underline{2.9} & \underline{1.3} & \textbf{1.4} & \textbf{0.7} & 2.6 & 1.3 & \textbf{2.2} & \textbf{1.0}\\
		\bottomrule
	\end{tabular}
\end{table*}

 \begin{table*}[t]
 \setlength\tabcolsep{3pt}
 \centering
 \caption{\textbf{Tracking performance on TUM RGB-D dynamic dataset}. The metric unit is [cm].}\label{tab2}
 	\begin{tabular}{lcccccccccccccccccc}
      \toprule
      \multirow{2}{*}{Method} & \multicolumn{2}{c}{\texttt{f3/sit\_s}} & \multicolumn{2}{c}{\texttt{f3/sit\_x}} & \multicolumn{2}{c}{\texttt{f3/sit\_r}} & \multicolumn{2}{c}{\texttt{f3/sit\_h}} & \multicolumn{2}{c}{\texttt{f3/walk\_s}} & \multicolumn{2}{c}{\texttt{f3/walk\_x}} & \multicolumn{2}{c}{\texttt{f3/walk\_r}} & \multicolumn{2}{c}{\texttt{f3/walk\_h}} & \multicolumn{2}{c}{Avg.}\\
      \cmidrule(lr){2-3}\cmidrule(lr){4-5}\cmidrule(lr){6-7}\cmidrule(lr){8-9}\cmidrule(lr){10-11}\cmidrule(lr){12-13}\cmidrule(lr){14-15}\cmidrule(lr){16-17}\cmidrule(lr){18-19}
       & RMSE & S.D. & RMSE & S.D. & RMSE & S.D. & RMSE & S.D. & RMSE & S.D. & RMSE & S.D. & RMSE & S.D. & RMSE & S.D. & RMSE & S.D.\\
      \midrule
      \rowcolor{gray!10}
      \multicolumn{19}{l}{\textit{Traditional}}\\
      % ORB-SLAM2 & 0.8 & 1.0 & 2.5 & 2.5 & 40.8 & 72.2 & 80.5 & 72.3 & 34.1\\
      % DSO & 1.7 & 11.5 & 3.7 & 12.4 & 1.5 & 12.9 & 13.8 & 40.7 & \\
      ORB-SLAM2 & {0.9} & \underline{0.4} & \underline{0.9} & \underline{0.5} & \underline{2.0} & \textbf{1.2} & {1.9} & 1.1 & 38.7 & 16.4 & 72.1 & 25.6 & 78.4 & 40.1 & 46.7 & 26.0 & 30.2 & 13.9 \\
      % DROID-SLAM & \textbf{0.5} & - & \underline{0.9} & - & \underline{2.2} & - & \underline{1.4} & - & 1.2 & - & 1.6 & - & 4.0 & - & 2.2 & -  & 1.8 & - \\
      ReFusion & {0.9} & - & 4.0 & -  & 13.2 & - & 11.0 & - & 1.7 & - & 9.9 & - & 40.6 & - & 10.4 & - & 11.5 & -\\
      DynaSLAM & \textbf{0.5} & \textbf{0.3} & 1.3 & 0.6 & 2.7 & \underline{1.4} & 1.9 & \underline{0.9} & {0.7} & \underline{0.3} & 1.6 & 0.9 & 3.5 & \textbf{1.9} & 3.0 & 1.6 & 1.9 & 1.0\\
      CFP-SLAM & \textbf{0.5} & \textbf{0.3} & \underline{0.9} & \textbf{0.4} & 2.5 & 1.5 & \underline{1.5} & \textbf{0.7} & {0.7} & \underline{0.3} & 1.4 & \underline{0.7} & 3.7 & 2.3 & 2.4 & 1.1 & 1.7 & \underline{0.9}\\
      %\hdashline
      \midrule
      \rowcolor{gray!10}\multicolumn{19}{l}{\textit{NeRF-based}}\\
      Co-SLAM & 1.5 & 0.8 & 4.8 & 2.4 & 51.7 & 41.7 & 8.6 & 4.3 & 30.6 & 9.3 & 72.5 & 33.6 & 236.0 & 31.5 & 100.7 & 31.8 & 63.3 & 19.4\\
      NID-SLAM & 1.9 & - & 7.5 & - & 8.6 & - & 10.5 & - & 6.2 & - & 6.4 & - & 64.8 & - & 7.1 & - & 14.1 & -\\
      RoDyn-SLAM & 1.7 & 0.9 & 5.2 & 3.1 & 6.7 & 3.9 & 4.4 & 2.2 & 1.7 & {0.9} & 8.3 & 5.5 & 7.9 & 4.7 & 5.6 & 2.8 & 5.2 & 3.0\\
%	         DDN-SLAM & - & - & 1.0 & - & - & - & 1.7 & - & 1.0 & - & \underline{1.3} & - & 3.9 & - & 2.3 & - & - & -\\
      DynaMoN & \textbf{0.5} & \textbf{0.3} & 1.0 & 0.6 & 2.4 & \underline{1.4} & 2.6 & 1.6 & {0.7} & \underline{0.3} & 1.4 & \underline{0.7} & 3.5 & \underline{2.1} & 1.9 & 0.9 & 1.8 & 1.0\\
      %\hdashline
      \midrule
      \rowcolor{gray!10}\multicolumn{19}{l}{\textit{3DGS-based}}\\
%	         SplaTAM & 1.4 & 0.7 & 1.6 & 1.0 & 12.8 & 5.9 & 13.0 & 6.5 & 105.0 & 28.2 & 139.8 & 39.0 & 119.4 & 40.3 & 113.0 & 55.6 & - & -\\
       SGS-SLAM & 1.4 & 0.7 & 1.8 & 1.1 & 13.5 & 8.1 & 12.9 & 5.9 & 60.3 & 10.2 & 130.4 & 35.3 & 70.7 & 26.5 & 69.6 & 37.9 & 37.2 & 15.7\\
      % MonoGS & 1.2 & 6.1 & 5.1 & 28.3 & 1.1 & 21.5 & 17.4 & 44.2 & \\
%	         Splat-SLAM & \textbf{0.5} & 0.3 & \underline{0.9} & 0.5 & 2.3 & 1.3 & 1.5 & 0.7 & 2.3 & 1.0 & \underline{1.3} & 0.6 & 4.5 & 3.1 & 2.0 & 1.1 & 1.9 & 1.1\\
      DG-SLAM & \underline{0.7} & 0.5 & 1.1 & \underline{0.5} & 3.6 & 2.2 & 2.5 & 1.1 & 3.5 & 2.5 & 2.9 & 1.5 & 4.4 & \underline{2.1} & 3.7 & 2.1 & 2.8 & 1.6\\
%	         PG-SLAM & 0.7 & 0.4 & 1.5 & 0.5 & 5.4 & 2.4 & 4.0 & 1.5 & 1.4 & 0.6 & 6.8 & 2.9 & - & - & 11.7 & 4.4 & - & -\\
      SDD-SLAM & - & - & - & - & - & - & - & - & \underline{0.5} & - & 1.5 & - & {3.2} & - & 2.0 & - & - & -\\
      WildGS-SLAM & \textbf{0.5} & \textbf{0.3} & \textbf{0.8} & \textbf{0.4} & 2.3 & \underline{1.4} & 1.7 & 1.1 & \underline{0.5} & \underline{0.3} & \underline{1.3} & \textbf{0.6} & \underline{3.1} & \underline{2.1} & \underline{1.5} & \underline{0.8} & \underline{1.5} & \underline{0.9}\\
      \textbf{Ours} & \textbf{0.5} & \textbf{0.3} & \textbf{0.8} & \textbf{0.4} & \textbf{1.9} & \textbf{1.2} & \textbf{1.3} & \textbf{0.7} & \textbf{0.4} & \textbf{0.2} & \textbf{1.2} & \textbf{0.6} & \textbf{2.9} & \textbf{1.9} & \textbf{1.4} & \textbf{0.7} & \textbf{1.3} & \textbf{0.8}\\
      \bottomrule
  \end{tabular}
\end{table*}

\section{EXPERIMENTS}
\subsection{Experimental Setup}
\textbf{Datasets.} We evaluate DL-SLAM on three dynamic datasets: the TUM RGB-D dynamic~\cite{tum} and BONN datasets~\cite{refusion}, as well as the challenging Wild-SLAM iPhone~\cite{wildgs} dataset to assess performance in unconstrained settings.

\noindent\textbf{Metrics.} We evaluate camera tracking accuracy using the Absolute Trajectory Error (ATE)~\cite{tum}, reporting both the Root Mean Square Error (RMSE) and Standard Deviation (S.D.). For reconstruction and rendering quality, we compare rendered images against the ground truth training views using Peak Signal to Noise Ratio (PSNR), SSIM~\cite{ssim}, and LPIPS~\cite{lpips} metrics. 

\noindent\textbf{Baselines.} We compare DL-SLAM against the state-of-the-art SLAM methods, which are divided into three groups based on their scene representation: $(1)$ Traditional: ORB-SLAM2~\cite{orb2}, ReFusion~\cite{refusion}, DynaSLAM~\cite{dyna}, CFP-SLAM~\cite{cfp}; $(2)$ NeRF-based: Co-SLAM~\cite{co}, NID-SLAM~\cite{nid}, RoDyn-SLAM~\cite{rodyn}, DynaMoN~\cite{dynamon}; $(3)$ 3DGS-based: SGS-SLAM~\cite{sgs}, DG-SLAM~\cite{dg}, SDD-SLAM~\cite{sdd}, WildGS-SLAM~\cite{wildgs}. For each category, we select one representative static method (ORB-SLAM2, Co-SLAM, and SGS-SLAM) as a reference, while the remaining are designed for dynamic environments.

\noindent\textbf{Implementation details.} We run DL-SLAM on a desktop PC with an Intel Core i9-12900KF CPU and an NVIDIA RTX 3090 GPU. For the cross-frame data association, the similarity threshold $\tau_\text{sim}$ is 0.7. The threshold $\tau_\text{prune}$ for dynamic Gaussian pruning is set to 0.8. For semantic refinement, we set $\tau_\text{match}=0.8$ and $\tau_\text{grad}=3$. For map optimization, the loss weights and variance threshold are set to $\lambda_\text{c}=0.5, \lambda_\text{reg}=10,  \lambda_\text{ssim}=0.2, \text{ and } \tau_\text{var}=0.002$, respectively.
%Additionally, we report the L1 distance between the rendered depth and the ground truth depth. %, following~\cite{nerf-slam}. 

% \begin{table}[t]
% % \setlength\tabcolsep{2pt}
% \centering
%     \caption{\textbf{Runtime and memory performance on TUM RGB-D} \texttt{f3/walk\_s}.}\label{tab4}
%     \begin{tabular}{lcccc}
	%         \toprule
	%       \multirow{2}*{Method} & Tracking & Mapping & Avg. Time & GPU\\
	%          & [ms/frame]$\downarrow$ & [ms/iter]$\downarrow$ & [ms/frame]$\downarrow$ & [MB]$\downarrow$\\
	%         \midrule
	%         Rodyn-SLAM & \underline{378.4} & 25.2 & 2723.6 & \textbf{5022}\\
	%         % Splat-SLAM & 7.20 & 9.89 & \textbf{23.95} & 10654\\ 
	%         % DG-SLAM & 89.2 & 13.7 & 645.9 & \underline{6954}\\
	%         SDD-SLAM & 557.4 & \textbf{9.3} & \underline{989.0} & \underline{7678}\\ 
	%         DL-SLAM(Ours) & \textbf{139.6} & \underline{11.6} & \textbf{661.6} & 11522\\ 
	%         \bottomrule
	%     \end{tabular}
% \end{table}

\subsection{Tracking Evaluation}
We report the camera tracking results in Table~\ref{tab1} and Table~\ref{tab2}. On the TUM RGB-D dataset, DL-SLAM consistently outperforms all baseline methods across every sequence. Notably, our approach achieves a significant improvement in average accuracy, with up to a 13\% reduction in RMSE compared to the next-best method. Similarly, on the BONN dataset, our method attains the best overall accuracy, surpassing competing approaches. These results validate that our dual-level probabilistic framework effectively mitigates the interference of dynamic objects, leading to robust camera tracking.

\subsection{Rendering Evaluation}
\begin{figure}[t]
	\centering
	\begin{minipage}[c]{0.32\linewidth}
		\centering\small\text{Before removal}
	\end{minipage}
	\begin{minipage}[c]{0.32\linewidth}
		\centering\small\text{After removal (RGB)}
	\end{minipage}
	\begin{minipage}[c]{0.32\linewidth}
		\centering\small\text{After removal (Sem)}
	\end{minipage}
	
	\centering
	\begin{minipage}[c]{0.32\linewidth}
		\centering
		\includegraphics[width=0.98\linewidth]{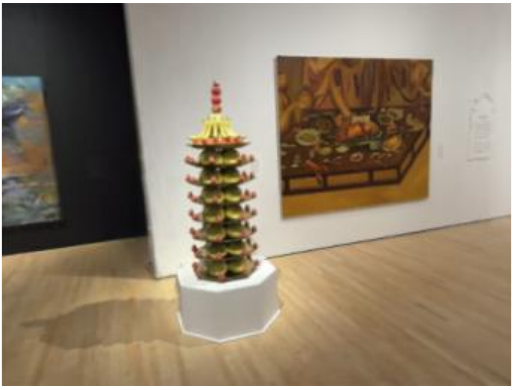}
	\end{minipage}
	\begin{minipage}[c]{0.32\linewidth}
		\centering
		\includegraphics[width=0.98\linewidth]{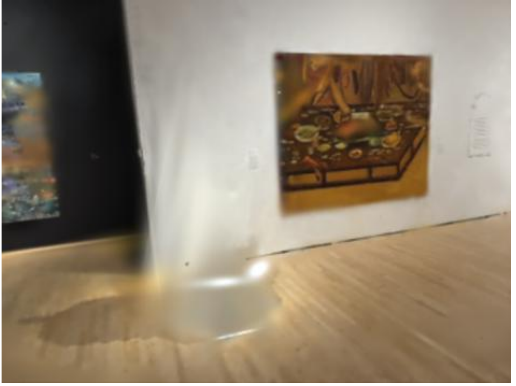}
	\end{minipage}
	\begin{minipage}[c]{0.32\linewidth}
		\centering
		\includegraphics[width=0.98\linewidth]{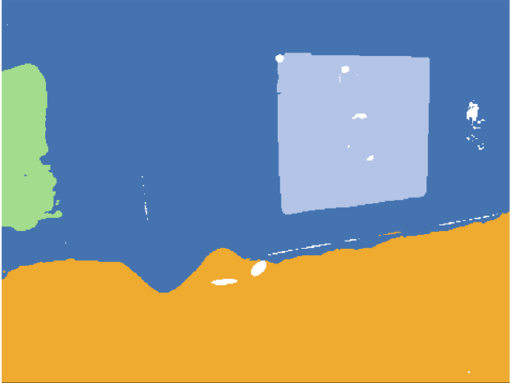}
	\end{minipage}
	
	\caption{\textbf{Interactive scene editing on Wild-SLAM iPhone}.}
	\label{fig5}
\end{figure}

Fig.~\ref{fig4} presents our rendering results on the Wild-SLAM iPhone and BONN datasets. In $\texttt{wandering}$ sequence, WildGS-SLAM exhibits rendering artifacts due to the integration of transiently static objects into its map. In contrast, our method yields artifact-free rendering by performing object-level pruning of dynamic objects. In the $\texttt{mv\_box}$ sequence, our method correctly treats the box that moves from the table to the floor as a dynamic object, removing it to avoid artifacts in either its initial or final location. Moreover, while the uncertainty estimation of WildGS-SLAM is inaccurate near dynamic object boundaries, our method estimates precise dynamic probability to ensure robust identification of dynamic regions. Table~\ref{tab3} presents the quantitative rendering results. Our method consistently outperforms competing methods, achieving photo-realistic rendering in challenging dynamic environments. In Fig.~\ref{fig5}, we demonstrate that our object-level semantic map supports interactive scene editing. By entering a user prompt specifying the target object, the corresponding instance is successfully identified and removed from the final map.

\begin{table}[t]
	\setlength\tabcolsep{3pt}
	\begin{center}
		\caption{\textbf{Runtime and memory performance on TUM RGB-D.}}\label{tab4}
		% \resizebox{\linewidth}{!}{
		\begin{tabular}{lcccc}
				\toprule
				\multirow{2}*{Metric} & Rodyn- & SGS- & WildGS- & \textbf{Ours}\\
				& SLAM & SLAM & SLAM & \\
				\midrule
				Segmentation [ms/frame]$\downarrow$ & \underline{250.0} & \textbf{168.5} & - & \textbf{168.5}\\
				Tracking [ms/frame]$\downarrow$ & \underline{378.4} & 1307.1 & \textbf{139.6} & \textbf{139.6}\\
				Mapping [ms/frame]$\downarrow$ & 757.2 & \underline{620.3} & 995.9 & \textbf{590.4}\\
%				Total Time [ms/frame]$\downarrow$ & 2723.6 &  & \textbf{649.5} & \underline{838.1}\\
				Model Size [MB]$\downarrow$ & \textbf{7.3} & 32.9 & \underline{8.0} & 11.1\\
				GPU Memory [GB]$\downarrow$ & \textbf{5.0} & \underline{6.5} & {10.8} & 11.5\\
				\bottomrule
		\end{tabular}
	\end{center}
\end{table}

\subsection{Runtime and Memory Analysis}
We report the runtime and memory usage of our method in Table~\ref{tab4}. For semantic segmentation, we provide SGS-SLAM with the same semantic labels generated by our method for a fair comparison. Although WildGS-SLAM bypasses this module, it still incurs substantial computational overhead from the online optimization of its uncertainty estimation network, which contributes to its higher mapping latency compared to ours. For tracking, our method shares the efficient DBA layer with WildGS-SLAM, achieving the lowest latency among all compared methods. In terms of memory, our method requires moderately higher GPU usage and model size than RoDyn-SLAM and WildGS-SLAM, a direct consequence of the additional semantic label and dynamic probability attributes stored per Gaussian. Compared to SGS-SLAM, which also maintains semantic attributes, our representation is approximately 3$\times$ more compact.

\subsection{Ablation Study}
\begin{figure}[t]
	\centering
	\begin{minipage}[c]{0.48\linewidth}
		\centering\small\text{Ground truth}
	\end{minipage}
	\begin{minipage}[c]{0.48\linewidth}
		\centering\small\text{w/o object-level}
	\end{minipage}
	
	\centering
	\begin{minipage}[c]{0.48\linewidth}
		\centering
		\includegraphics[width=0.98\linewidth]{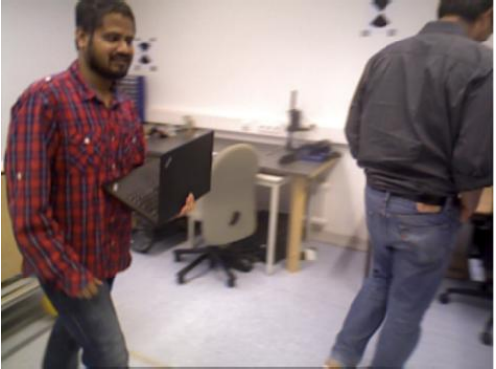}
	\end{minipage}
	\begin{minipage}[c]{0.48\linewidth}
		\centering
		\includegraphics[width=0.98\linewidth]{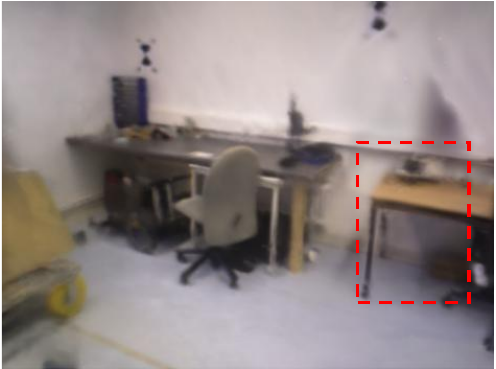}
	\end{minipage}
	
	\begin{minipage}[c]{0.48\linewidth}
		\centering\small\text{DL-SLAM (Ours)}
	\end{minipage}
	\begin{minipage}[c]{0.48\linewidth}
		\centering\small\text{Zoomed-in view}
	\end{minipage}
	
	\begin{minipage}[c]{0.48\linewidth}
		\centering
		\includegraphics[width=0.98\linewidth]{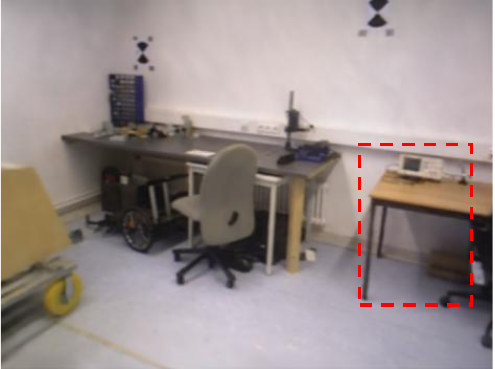}
	\end{minipage}
	\begin{minipage}[c]{0.48\linewidth}
		\centering
		\includegraphics[width=0.98\linewidth]{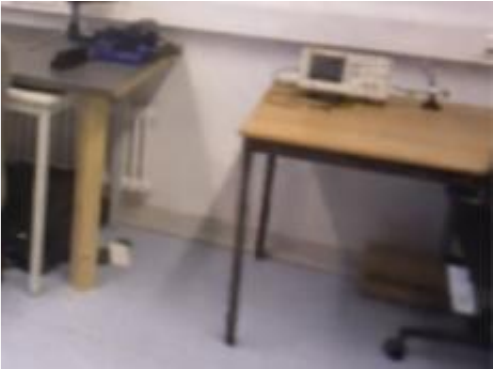}
	\end{minipage}
	
	\caption{\textbf{Rendering ablation results on BONN}.}
	\label{fig6}
\end{figure}

We conduct an ablation study on the BONN dataset to validate the effectiveness of our core components. 
%w/o pixel-level prob. refers to disable the entire dynamic probability framework, as the object-level probability is dependent on the initial pixel-level estimates. w/o object-level prob. refers to disable object-level pruning of dynamic Gaussians. 
As shown in Table~\ref{tab5}, the pixel-level probability proves crucial for robust tracking, as it mitigates interference from dynamic objects. As illustrated in Fig.~\ref{fig6}, the object-level probability enables identifying and pruning transiently static objects, effectively eliminating rendering artifacts caused by dynamic objects. We also ablate our semantic refinement strategy to demonstrate its crucial role in enhancing rendering quality and enabling the generation of high-fidelity semantic maps.

\begin{table}[t]
	\centering
	\caption{\textbf{Ablation study on BONN dataset}. We report the average performance across 8 sequences. The units for ATE
		RMSE, ATE S.D. and PSNR are [cm], [cm], and dB.}\label{tab5}
	\begin{tabular}{lccc}
		\toprule
		& ATE RMSE$\downarrow$ & ATE S.D.$\downarrow$ & PSNR$\uparrow$\\
		\midrule
		w/o pixel-level prob. & 4.8 & 2.8 & 19.00\\
		w/o object-level prob. & 2.6 & 1.3 & \underline{19.38}\\
		w/o semantic refinement & \underline{2.4} & \underline{1.1} & 18.74\\
		DL-SLAM (Ours) & \textbf{2.2} & \textbf{1.0} & \textbf{19.99}\\
		\bottomrule
	\end{tabular}
\end{table}

\section{Conclusion}
In this paper, we present DL-SLAM, a high-fidelity Gaussian Splatting SLAM system in dynamic environments. At the core of our method is a dual-level probabilistic framework that resolves the challenge of handling transiently static objects. By coupling pixel-level estimation with object-level understanding, our method successfully leverages these objects to enhance tracking accuracy while simultaneously ensuring their exclusion from the final map. Furthermore, our dynamic-aware semantic refinement strategy compensates for temporal inconsistencies and occlusions caused by moving objects. As validated through extensive experiments, DL-SLAM outperforms existing methods in terms of tracking accuracy and rendering quality. A promising direction for future work is to extend our framework to explicitly model the trajectories of dynamic objects, enabling a complete 4D scene understanding.

\bibliographystyle{ACM-Reference-Format}
\bibliography{root}

\appendix

\end{document}